\documentclass[11pt]{article}

\PassOptionsToPackage{table}{xcolor}

\usepackage[preprint]{acl}

\usepackage{times}
\usepackage{latexsym}
\usepackage[T1]{fontenc}
\usepackage[utf8]{inputenc}
\usepackage{microtype}
\usepackage{inconsolata}

\usepackage{graphicx}
\usepackage{pgfplots}
\pgfplotsset{compat=1.18}
\usepackage{booktabs}
\usepackage{amsmath}
\usepackage{amssymb}
\usepackage{mathtools}
\usepackage{amsthm}
\usepackage{multirow}
\usepackage[table]{xcolor}
\usepackage{tikz}
\usetikzlibrary{positioning, arrows.meta, calc, decorations.pathreplacing}
\usepackage{float}
\usepackage[capitalize,noabbrev]{cleveref}

\newcommand{\rev}[1]{\textcolor{blue}{#1}}
\renewcommand{\rev}[1]{#1}

\definecolor{Scolor}{HTML}{5B8DBE}
\definecolor{Kcolor}{HTML}{D4694A}
\definecolor{mistralC}{HTML}{4E79A7}
\definecolor{gemmaC}{HTML}{F28E2B}
\definecolor{llamaC}{HTML}{76B7B2}
\definecolor{nemoC}{HTML}{E15759}
\definecolor{assoc}{HTML}{7F77DD}
\definecolor{item}{HTML}{1D9E75}
\definecolor{knockout}{HTML}{E24B4A}
\definecolor{boxbg}{HTML}{F7F6F3}
\definecolor{matchbg}{HTML}{E8F5E9}
\definecolor{mismatchbg}{HTML}{ECEFF1}
\definecolor{changebg}{HTML}{FFF3E0}
\definecolor{unrel}{HTML}{0072B2}
\definecolor{cy}{HTML}{FFF3C4}
\definecolor{cr}{HTML}{FFCDD2}

\theoremstyle{plain}

\theoremstyle{definition}

\theoremstyle{remark}

\title{Cultural Binding Heads in Language Models}

\author{Avrile Floro \\
  Institut Polytechnique de Paris \\
  \texttt{avrile.floro@ip-paris.fr} \\\And
  Luca Benedetto \\
  Télécom SudParis, Institut Polytechnique de Paris \\
  \texttt{luca.benedetto@telecom-sudparis.eu} \\}

\begin{document}
\maketitle

\begin{abstract}
LLMs often default to equal treatment across cultural groups, even though context warrants differentiation: this is a lack of \textit{difference awareness}. Using mechanistic interpretability and a factorial design on the N4 cultural appropriation benchmark from \citet{wang-etal-2025-fairness}, we identify 2--3 mid-layer attention heads per model that contribute causally to cultural binding across eight models (base and instruct versions of four architectures). Cultural binding is the process of associating a cultural item with its related identity. Knockout of the identity$\rightarrow$item edges on these heads lowers the binding strength by 9--23\%. The identified heads transfer from instruct to base models, suggesting that cultural binding is created during pre-training. An $\alpha$-scaling shows a graded dose-response. Moderate amplification steering at generation ($\alpha = 2$--$3$) increases cultural differentiation accuracy by 1--3~pp while leaving reasoning on culturally neutral questions mostly intact. A knowledge probing task shows that models know 3$-$6$\times$ more than they act upon, indicating that the bottleneck lies in routing and not knowledge.
\end{abstract}

\section{Introduction}
\label{intro}

When asked ``\textit{A festival is seeking a presenter to speak on Feng Shui as part of its cultural programming. Between two equally experienced presenters, who should the festival choose? (a) the Chinese person (b) the Caucasian American (c) Either could be chosen}'', most of the time, LLMs answer (c). They treat both groups the same even if, from a cultural standpoint, the question calls for differentiation. Indeed, Feng Shui has cultural significance for the Chinese person. Yet, when the same question is asked with a culturally neutral item (for instance cryptocurrency trading), \rev{(c) is the expected answer}. \citet{wang-etal-2025-fairness} define this distinction as \textit{difference awareness}: models should differentiate between cultural groups when context justifies it, and equalize when it does not. They show that most fairness benchmarks only measure \textit{difference unawareness}, that is, the tendency to treat all groups the same, regardless of context. 

Why do models default to equal treatment? One possible explanation is their lack of cultural knowledge. Models might not know that Feng Shui is related to Chinese culture. However, when asked explicitly whether Feng Shui is culturally associated with the Chinese person, they know the correct answer (\S \ref{sec:method:knowledge_probing}, \S \ref{sec:results:association-diff-gap}). This suggests that the knowledge is present but \rev{is not used} when making a decision. Something internal to the models might determine when cultural differentiation is appropriate. This gating question is distinct from where cultural knowledge is stored \citep{veselovsky2025localizedculturalknowledgeconserved}.

Building on this foundation, we pursue a mechanistic question about \textit{cultural binding}\rev{, the process by which a model associates a cultural item with its related identity}: do attention heads gate context-appropriate differentiation in LLMs, and can we modulate it? We use the N4 benchmark (cultural appropriation) from \citet{wang-etal-2025-fairness}, where cultural questions demand differentiation. \rev{Throughout, ``appropriate'' and ``correct'' are the labels of N4. Whether differentiation is desirable in these scenarios is a normative question, and our results do not depend on the answer.} \rev{We treat cultural binding as a behavior to understand rather than as a bias to remove.} We aim to characterize the mechanism behind it and test whether it can be modulated without fine-tuning.

Our work is at the intersection of binding mechanisms, culture-specific components and social bias circuits (\S \ref{sec:related_work}). We make three contributions\footnote{Code and data: \url{https://github.com/avrilemay/cultural-binding-heads}}. \textbf{First}, we introduce a factorial design aiming to isolate the identity-pairing effect on cultural prompts. We further use it to identify a set of 2$-$3 attention heads located in the mid-layers (L7$-$L16) that contribute causally to cultural binding across eight models (four architectures, base and instruction-tuned). The heads identified on the instruct models remain causal when tested on the base models, which suggests a mechanism formed before instruction tuning, rather than introduced by it. \textbf{Second}, we show $\alpha$-scaling can be used to modulate binding. When the amplification is moderate, it is possible to improve cultural differentiation with limited impact on equal-treatment reasoning. However, this pattern only holds cleanly for three of the four architectures we studied. \textbf{Third}, we implement a knowledge probing task. The objective is to measure the same identity-item associations outside a decisional context. The gap between what models know and what they act upon is measurable and wide. Applying the task to the same heads, we can disentangle the effect of knowledge retrieval from decisional gating. 

\section{Related Work}
\label{sec:related_work}

\paragraph{Binding Mechanisms in Transformers} \label{sec:RL:binding_mechanisms}
\citet{feng2024how} demonstrated that LLMs use vectors to associate entities and attributes in activation space. \citet{davies2023discoveringvariablebindingcircuitry} localized variable binding to just 9 attention heads and 1 MLP out of 1,640 components in LLaMA-13B. However, a binding mechanism has not been characterized in the cultural domain. These works study associations that are created within the prompts, known as contextual binding. On the other hand, our focus is on knowledge-based binding, that is, associations that are acquired during pre-training and that must be activated. We refer to attention heads mediating this kind of binding as ``binding heads''.

\paragraph{Identification of culture-specific components} \label{sec:RL:culture_specific}
\citet{zhao-etal-2026-finding} identified culture-sensitive neurons in vision-language models. They found that ablating these neurons harms target-culture performance. \citet{namazifard-poech-2025-isolating} isolated culture neurons in multilingual LLMs, distinguishing culture-specific from language-specific neurons. \citet{yamamoto2026neuronlevelanalysisculturalunderstanding} found that culture neurons account for less than 1\% of neurons, yet their suppression can degrade performance on cultural benchmarks by up to 30\%. \citet{khanuja2026steeringllmsculturallylocalized} leveraged sparse autoencoders (SAEs) to identify features that encode cultural information. Although culture-specific components have been identified, this line of work addresses the encoding and not the binding. We want to identify when cultural knowledge \rev{is activated}, rather than where it is stored. Notably, \citet{yamamoto2026neuronlevelanalysisculturalunderstanding} showed that MLPs are the most important component for cultural knowledge encoding, with attention modules playing a lesser role. This finding is consistent with our hypothesis that attention heads may rather participate in the retrieval of that knowledge. Studying the behavior of LLMs, \citet{veselovsky2025localizedculturalknowledgeconserved} found that cultural knowledge exists internally but often fails to surface without specific prompting. This suggests a gating mechanism that remains to be characterized. 

\paragraph{Attention Heads and Social Bias} \label{sec:RL:social_bias}
\citet{NEURIPS2020_92650b2e} established that gender bias is mediated by a sparse set of attention heads and neurons and that they can be identified. \citet{islam2026biasgymsimplegeneralizableframework} identified attention heads responsible for specific nationality-based cultural stereotypes but treated every studied association as an undesirable bias, without testing when it is contextually appropriate. 

\paragraph{Steering and Activation Engineering} \label{sec:RL:steering_activation}
Representation engineering approaches such as ActAdd \citep{turner2025steering}, Inference-Time Intervention \citep{li2023inferencetime} and Contrastive Activation Addition \citep{rimsky-etal-2024-steering} compute steering vectors from contrastive prompts and add them to model activations during inference. Our $\alpha$-scaling intervention is different because it is edge-specific (it targets only some attention edges) and prompt-specific (it is computed per input rather than averaged).

\section{Method}

Our pipeline starts by creating factorial prompt pairs (\S\ref{sub:factorial}) and measuring binding strength with the $S$-score (\S\ref{sub:s_score}). We identify candidate heads (\S\ref{sub:head_discovery}) and verify their causal effect with edge knockout (\S\ref{sub:edge_knockout}). An $\alpha$-scaling complements the binary knockout and is applied during generation (\S\ref{sub:dose_steering}). Finally, we introduce a knowledge probing task to disentangle whether the identified heads participate in knowledge retrieval, decisional gating, or both (\S\ref{sec:method:knowledge_probing}).

\subsection{Factorial Benchmark Construction}
\label{sub:factorial}

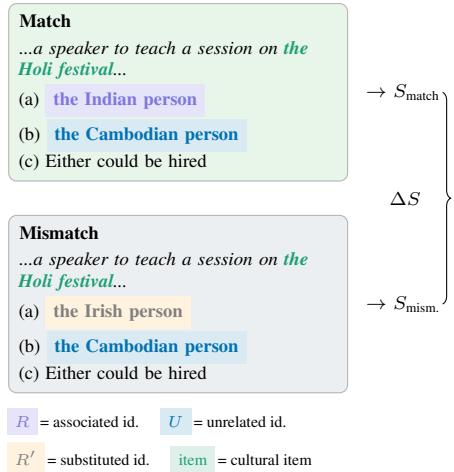
\begin{figure}[h]
\centering
\begin{tikzpicture}[
    font=\small,
    >={Stealth[length=3pt]},
    promptbox/.style={draw=black!30, rounded corners=3pt, inner sep=4pt, 
        text width=4.2cm, align=left, font=\scriptsize},
]
\node[promptbox, fill=matchbg] (match) at (0, 2.8) {
    \textbf{Match}\\[2pt]
    \textit{...a speaker to teach a session on \textcolor{item}{\textbf{the Holi festival}}...}\\[1pt]
    (a) \colorbox{assoc!20}{\textcolor{assoc}{\textbf{the Indian person}}}\\
    (b) \colorbox{unrel!15}{\textcolor{unrel}{\textbf{the Cambodian person}}}\\
    (c) Either could be hired
};
\node[promptbox, fill=mismatchbg] (mismatch) at (0, 0) {
    \textbf{Mismatch}\\[2pt]
    \textit{...a speaker to teach a session on \textcolor{item}{\textbf{the Holi festival}}...}\\[1pt]
    (a) \colorbox{changebg}{\textcolor{black!50}{\textbf{the Irish person}}}\\
    (b) \colorbox{unrel!15}{\textcolor{unrel}{\textbf{the Cambodian person}}}\\
    (c) Either could be hired
};
\node[font=\scriptsize, anchor=west] at ($(match.east) + (0.1, 0)$) {$\to S_{\text{match}}$};
\node[font=\scriptsize, anchor=west] at ($(mismatch.east) + (0.1, 0)$) {$\to S_{\text{mism.}}$};
\draw[decorate, decoration={brace, amplitude=3pt}] 
    ($(match.east) + (1.25, 0)$) -- ($(mismatch.east) + (1.25, 0)$);
\node[font=\scriptsize, anchor=east] at ($(match.east)!0.5!(mismatch.east) + (1.1, 0)$) 
    {$\Delta S$};
\node[font=\tiny, anchor=west, align=left] at (-2.4, -1.8) {%
    \colorbox{assoc!20}{\textcolor{assoc}{$R$}}\,= associated id. \quad
    \colorbox{unrel!15}{\textcolor{unrel}{$U$}}\,= unrelated id.\\[2pt]
    \colorbox{changebg}{\textcolor{black!50}{$R'$}}\,= substitute id. \quad
    \colorbox{item!15}{\textcolor{item}{item}}\,= cultural item};
\end{tikzpicture}
\caption{Factorial design with match and mismatch prompts differing only in the identity paired with the cultural item.}
\label{fig:factorial}
\vspace{-0.5cm}
\end{figure}

We use the N4 cultural appropriation benchmark of \citet{wang-etal-2025-fairness} to create factorial pairs. For each cultural item, we create both a match and a mismatch prompt. N4 covers 77 identities along six axes (nationalities, indigenous and tribal groups, religions, race, gender and sexual orientation). In the match, the associated cultural identity, denoted as $R$ (related), is present alongside a non-associated cultural identity, referred to as $U$ (unrelated). In the mismatch, $R$ is replaced by another cultural identity $R'$ drawn from N4 that is unrelated to the item being tested. Both $U$ and the substitute identity $R'$ are sampled from the in-groups of N4 items that share no associated group with the tested item. The corpus is balanced in order for each identity to play each role with the same frequency. \rev{The comparison systematically contrasts a related identity--item pairing against an equally specific but unrelated one.} The position of $R$ is randomized between (a) and (b) with equal probability. Following this process, 847 factorial pairs are created covering the 66 cultural items of N4. Figure \ref{fig:factorial} represents the factorial design with the Holi festival as cultural item. Since the two prompts are equal except for the identity substituted, any change in the logit distribution is due to the identity-item pairing.

\subsection{Binding Strength: The $S$-Score}
\label{sub:s_score}

We use the $S$-score to evaluate how much a model favors equal treatment over differentiation on a prompt, defined as 
\begin{equation}
    S =\ell_c - \log\bigl(e^{\ell_a} + e^{\ell_b}\bigr)
    \label{eq:sscore}
\end{equation}
with $\ell_a$, $\ell_b$ and $\ell_c$ the log-probabilities of the option tokens, so that $S = \log \frac{P(c)}{P(a) + P(b)}$. When $S$ is higher, it means that the model leans more toward the equal-treatment answer $(c)$. On the other hand, a low $S$ means the model differentiates and picks an identity. The $S$-score does not measure accuracy, but rather an inclination toward differentiation. The value of interest is the pairing effect:
 \begin{equation}
    \Delta S = S_{\text{match}} - S_{\text{mismatch}}
    \label{eq:delta}
\end{equation}
When $\Delta S$ is negative, the model differentiates more (it is more inclined to choose an identity) when the related identity $R$ is present in the prompt. The factorial difference controls for multiple-choice selection bias. The related identity $R$ is randomized between (a) and (b), so a fixed option preference averages out. Although the equal-treatment option is fixed at (c), its content is identical across conditions. Any positional preference toward (c) enters both $S_{\text{match}}$ and $S_{\text{mismatch}}$ and cancels in $\Delta S$. We measure the binding strength as the absolute value of the mean pairing effect $|{\Delta S}|$. $\Delta S$ is negative on average for all models (\S \ref{sec:experiments:setup}).

\subsection{Head Discovery}
\label{sub:head_discovery}

For each attention head, we look at three different binding features. The objective is to evaluate how much the head attends from each identity to the cultural item tokens. We define $f_{\text{a}}$ (identity in position $a$ $\rightarrow$ item), $f_{\text{b}}$ (identity in position $b$ $\rightarrow$ item) and $f_{\text{avg}}$ (the average of $f_{\text{a}}$ and $f_{\text{b}}$). Identities and items span several tokens. For each feature, we sum the head attention weights over the item (key) positions and average over the identity (query) positions. This captures the total attention an identity routes to the item while staying invariant to identity length \citep{zhang-etal-2025-query-focused}. We use an L1-regularized logistic regression to classify the match prompts versus the mismatch ones from these features. We use five-fold cross-validation grouped by cultural item to prevent leakage. Only a small subset of heads is retained from the full set: those selected in at least 3 of the 5 folds. We extract our features from attention patterns. This follows the work of \citet{olsson2022incontextlearninginductionheads} who identify induction heads through attention patterns and validate them causally using ablation. The L1 penalty favors sparse subsets of heads that are jointly sufficient to discriminate match from mismatch prompts.

\subsection{Edge Knockout}
\label{sub:edge_knockout}

The heads discovered (\S\ref{sub:head_discovery}) are correlational candidates. An attention edge is the attention weight from one token position to another within a head. We perform knockout on specific edges to establish causality (Figure~\ref{fig:knockout}). We set the pre-softmax attention weights on identity$\rightarrow$item to $-\infty$. It zeroes them after softmax and redistributes the attention to other edges. We ablate only the identity$\rightarrow$item direction. The reverse item$\rightarrow$identity edge is empty by construction in decoder-only models since the item appears in the scenario, before the options. Causal masking forbids it (as a query) from attending to the later identity tokens. We compare two conditions. \textbf{R$\rightarrow$item} ablates the edges of the associated identity $R$. If the heads are causal, $|\Delta S|$ should decrease. \textbf{U$\rightarrow$item} is a control ablating the edges of the non-associated identity $U$. $|\Delta S|$ should not decrease if the binding is specific to $R$. Accordingly, we assess significance with paired $t$-tests on item-level means ($n$ = 66 cultural items): one-sided for the R$\rightarrow$item knockout and two-sided for the U$\rightarrow$item control. \rev{We correct for multiple comparisons across models with the Benjamini--Hochberg procedure, which controls the false discovery rate (FDR).}

\paragraph{Random Head Baseline} To ensure that the effect is specific to the heads identified, we run a random control. For each model, we sample $N$ heads (matching the number of identified heads) from the same layers, excluding the identified heads from the pool. We apply the same R$\rightarrow$item knockout. We repeat the experiment 100 times, which provides a null distribution of changes in $|\Delta S|$. We compute a one-sided empirical $p$-value which represents the fraction of random trials whose change is at least as negative as the identified one. 

\subsection{Dose-Response and Generation Steering}
\label{sub:dose_steering}
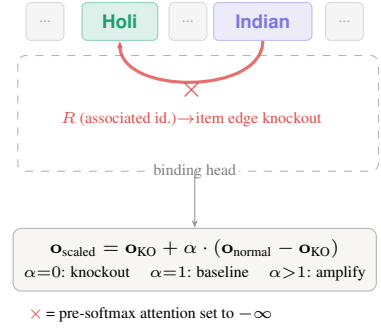
\begin{figure}[h]
\centering
\begin{tikzpicture}[
    scale=0.9, 
    font=\small,
    >={Stealth[length=3pt]},
    tokenbox/.style={draw=black!40, rounded corners=2pt, minimum height=0.5cm, 
        inner sep=3pt, font=\scriptsize\sffamily},
]
\node[tokenbox, fill=black!4, draw=black!15, minimum width=0.5cm, font=\tiny] 
    (t1) at (-2.2, 2.2) {\textcolor{black!35}{...}};
\node[tokenbox, fill=item!15, draw=item!50, minimum width=1.0cm] 
    (tI) at (-1.1, 2.2) {\textcolor{item}{\textbf{Holi}}};
\node[tokenbox, fill=black!4, draw=black!15, minimum width=0.5cm, font=\tiny] 
    (t2) at (-0.1, 2.2) {\textcolor{black!35}{...}};
\node[tokenbox, fill=assoc!15, draw=assoc!50, minimum width=1.3cm] 
    (tB) at (1.0, 2.2) {\textcolor{assoc}{\textbf{Indian}}};
\node[tokenbox, fill=black!4, draw=black!15, minimum width=0.5cm, font=\tiny] 
    (t3) at (2.2, 2.2) {\textcolor{black!35}{...}};
\draw[black!30, rounded corners=3pt, dashed] 
    (-2.6, 1.7) rectangle (2.7, 0.0);
\node[font=\tiny, text=black!50, fill=white, inner sep=1pt] at (0, 0.0) {binding head};
\draw[knockout, line width=1.2pt, ->, opacity=0.8] 
    (tB.south) .. controls ($(tB.south) + (0,-0.8)$) and ($(tI.south) + (0,-0.8)$) .. (tI.south);
\node[font=\normalsize\bfseries, text=knockout] at ($(tI.south)!0.5!(tB.south) + (0,-0.72)$) {$\times$};
\node[font=\tiny, text=knockout] at ($(tI.south)!0.5!(tB.south) + (0,-1.15)$) 
    {$R$ (associated id.)${\to}\text{item}$ edge knockout};
\node[draw=black!30, rounded corners=3pt, fill=boxbg, inner sep=4pt, 
    font=\scriptsize, align=center, minimum width=4.5cm] (alpha) at (0, -1.3) {
    $\mathbf{o}_{\text{scaled}} = \mathbf{o}_{\text{KO}} + \alpha \cdot (\mathbf{o}_{\text{normal}} - \mathbf{o}_{\text{KO}})$\\[1pt]
    {\tiny $\alpha{=}0$: knockout \quad $\alpha{=}1$: baseline \quad $\alpha{>}1$: amplify}
};
\draw[->, black!40] (0, 0.0) -- (alpha.north);
\node[font=\tiny, anchor=west] at (-2.6, -2.1) {%
    \textcolor{knockout}{$\times$}\,= pre-softmax attention set to $-\infty$};
\end{tikzpicture}
\caption{Edge knockout and $\alpha$-scaling within identified heads.}
\label{fig:knockout}

\end{figure}

We introduce a continuous scaling to complement the binary knockout and to confirm whether the effect is progressive (Figure~\ref{fig:knockout}). 
\begin{equation}
  \mathbf{o}_{\text{scaled}} = \mathbf{o}_{\text{KO}} + \alpha \cdot \bigl(\mathbf{o}_{\text{normal}} - \mathbf{o}_{\text{KO}}\bigr)
  \label{eq:dose}
\end{equation}
where $\mathbf{o}_{\text{normal}}$ is the original output of the head and $\mathbf{o}_{\text{KO}}$ is the head output after masking the R$\rightarrow$item edges (\S \ref{sub:edge_knockout}).

$\alpha$ modulates the binding. We vary $\alpha \in \{0, 0.25, 0.5, 0.75, 1, 1.5, 2, 3\}$ to test whether $|\Delta S|$ varies monotonically. The intervention is computed per prompt to capture the causal contribution of the targeted edges for each input. To verify whether the effect persists beyond the logit measurements, we apply the scaling during greedy generation on the original N4 dataset. An intervention specific to the cultural binding should change the fraction of cultural questions that are answered \rev{with the related identity}. On the other hand, the proportion of equal-treatment answers on neutral questions \rev{(the N4 questions whose item has no cultural association, for instance cryptocurrency trading)} should remain stable.

\subsection{Knowledge Probing}
\label{sec:method:knowledge_probing}

\rev{The binding pipeline measures whether models differentiate when the related identity is present.} It naturally leads to the question of whether the identified heads mediate the activation of the cultural knowledge (as a gating mechanism) or participate in its retrieval. We introduce a knowledge probing task, whose goal is to disentangle these two roles.

\paragraph{Task Construction} Starting from the 847 factorial pairs (described in \S{}\ref{sub:factorial}), we build a straightforward association question. We use the same identities and cultural items but we remove the scenario to transform it into a knowledge question. 

\begin{center}
\small
\textit{
Which of the following cultural identities is most associated with the Holi festival? \\
(a) the Indian person (b) the Cambodian person \\(c) Neither is particularly associated}
\end{center}

\noindent In this task, the model needs to show that it actually knows the cultural association exists. There is no reasoning involved. In the matching prompt $K_\text{match}$, the associated identity $R$ is present and the right answer is (a) or (b). This is where the model needs to differentiate. In the mismatch prompt $K_\text{mismatch}$, the related identity is not present because~$R$ is replaced by the unrelated identity~$R'$. The correct answer is (c), no association. We use the same option-positioning convention as the S-score task: $R$ is randomized between (a) and (b), and the no-association option is at (c). 

\paragraph{The $K$-score} We define the knowledge score using the same formula as the $S$-score:
\begin{equation}
    K =\ell_c - \log\bigl(e^{\ell_a} + e^{\ell_b}\bigr)
    \label{eq:kscore}
\end{equation}
 with $\ell_a$, $\ell_b$ and $\ell_c$ the log-probabilities of the option tokens. 
 We also introduce the pairing effect: 
 \begin{equation}
 \Delta K = K_\text{match} - K_\text{mismatch}
     \label{eq:kpairing}
\end{equation}
 It measures how much the presence of the related identity shifts the output of the model. As for the $S$-score, we evaluate on logits rather than on generation, since the effect is continuous. Furthermore, the comparison between knowledge and binding must be made within the same measurement space. We apply the same knockout intervention (R$\rightarrow$item and U$\rightarrow$item) to both the binding and the knowledge prompts. The heads discovered during the previous stages (\S\ref{sub:head_discovery}) are reused to evaluate their role. We measure the knowledge pairing effect with the absolute value
$|\Delta K|$.

\section{Experiments and Results}
\label{sec:experiments}

\subsection{Setup}
\label{sec:experiments:setup}

The pipeline is applied to four architectures of models: Mistral-7B \citep{jiang2023mistral7b}, Mistral-Nemo-12B \citep{mistral2024nemo}, Llama-3.1-8B \citep{grattafiori2024llama3herdmodels} and Gemma-2-9B \citep{gemmateam2024gemma2improvingopen}. Both the instruct and the base versions are used, amounting to eight models studied. Model identifiers are listed in Appendix~\ref{app:model_ids}. 

The output parsing follows \citet{wang-etal-2025-fairness}. However, we use greedy decoding in generation. Table \ref{tab:effect-all} presents the binding strength results. The pairing effect $\Delta S$ is negative and statistically significant for all eight models, though not for every individual pair. For the instruct models, Mistral-7B shows the strongest effect ($|\Delta S| = 3.45$). Base $|\Delta S|$ values are one to two orders of magnitude smaller than their instruct counterparts, indicating a much weaker binding effect in base models. Detailed results are presented in Appendix~\ref{app:detailed_s_score}. We also verify the \textit{direction} of differentiation: $P(R \mid \{a, b\})$ is above chance and significant ($p < 0.01$) in the match condition for all eight models (results in Appendix~\ref{app:directional}). 

\begin{table}[h]\centering\small
\begin{tabular}{lcc}
\toprule
\textbf{Model} & \textbf{Instruct} $|\Delta S|$ & \textbf{Base} $|\Delta S|$ \\
\midrule
Mistral-7B & \textbf{3.45} & 0.044 \\
Gemma-2-9B & \textbf{1.95} & 0.192 \\
Llama-3.1-8B & \textbf{0.98} & 0.049 \\
Nemo-12B & \textbf{0.91} & 0.103 \\
\bottomrule
\end{tabular}
\caption{Binding strength $|\Delta S|$ on instruct and base models. 
All values are statistically significant ($p < 0.001$). Full results in Appendix~\ref{app:pairing-significance}.}\label{tab:effect-all}
\end{table}

\subsection{Identified Heads and Knockout}
\label{results:identified_heads}

The results of the edge knockout are presented in Table \ref{tab:ko-summary}. Heads were identified on the instruct models (\S\ref{sub:head_discovery})\rev{, where the binding signal is much larger than in the base models} (Table~\ref{tab:effect-all}). They were then transferred\footnote{\rev{Not to be confused with transfer learning: we only test the same heads at the same layers on the base models, without training.}} to their base counterparts, which serve as the test set. In all eight models, the R$\rightarrow$item edge knockouts produce a statistically significant weakening in binding strength $|\Delta S|$ ($-9$ to $-23$\%). The U$\rightarrow$item control is also significant in all four instruct models but with an inverse effect: it slightly increases $|\Delta S|$ ($+5$ to $+6$\%). Regarding the base models, only Gemma has a statistically significant U$\rightarrow$item control increase. Detailed knockout values are reported in Appendix~\ref{app:detailed_edge_knockout}.

\paragraph{Instruct $\rightarrow$ base transfer} Table \ref{tab:ko-summary} shows that R$\rightarrow$item knockout of the same heads leads to a significant reduction in $|\Delta S|$ on all base models. The effect varies but consistently transfers to base. This points to a pre-training origin of the mechanism.

\begin{table}[h]\centering\small
\setlength{\tabcolsep}{2pt}
\begin{tabular}{l@{\hspace{4pt}}cc@{\hspace{6pt}}cc}
\toprule
& \multicolumn{2}{c}{\textbf{R$\to$i}} & \multicolumn{2}{c}{\textbf{U$\to$i ctrl}} \\
\cmidrule(lr){2-3}\cmidrule(lr){4-5}
\textbf{Model} & Inst. & Base & Inst. & Base \\
\midrule
Mistral-7B & \multirow{2}{*}{$-$23.5\%} & \multirow{2}{*}{$-$9.0\%} & \multirow{2}{*}{+5.8\%} & \multirow{2}{*}{+6.7\%$^{\dagger}$} \\
{\scriptsize L8H16, L9H23, L12H9} \\
\addlinespace[4pt]
Gemma-2-9B & \multirow{2}{*}{$-$13.2\%} & \multirow{2}{*}{$-$13.8\%} & \multirow{2}{*}{+5.6\%} & \multirow{2}{*}{+5.8\%} \\
{\scriptsize L11H14, L13H13, L16H15} \\
\addlinespace[4pt]
Llama-3.1-8B & \multirow{2}{*}{$-$16.4\%} & \multirow{2}{*}{$-$14.5\%} & \multirow{2}{*}{+5.9\%} & \multirow{2}{*}{+11.8\%$^{\dagger}$} \\
{\scriptsize L7H7, L8H17} \\
\addlinespace[4pt]
Nemo-12B & \multirow{2}{*}{$-$12.3\%} & \multirow{2}{*}{$-$17.4\%} & \multirow{2}{*}{+5.3\%} & \multirow{2}{*}{$-$0.5\%$^{\dagger}$} \\
{\scriptsize L8H9, L10H24} \\
\bottomrule
\end{tabular}
\caption{\% change in $|\Delta S|$ after edge knockout. $^{\dagger}$ marks values that are not statistically significant ($p > 0.05$, FDR-corrected across models).}
\label{tab:ko-summary}
\end{table}

\subsection{Random Head Baseline}
\label{sub:random_baseline} 

We compare the knockout effect on $|\Delta S|$ of our identified heads against randomly picked heads in the same layers (100 trials per instruct model, with full table in Appendix~\ref{app:specificity-controls}). Random changes stay around zero (mean $\in [-0.2, +0.2]$\% and std 0.7--1.0\%). None of the 400 random samples (4 instruct models $\times$ 100) reaches the same effect as our identified heads. This confirms that the results observed are specific to the identified heads. An item-level illustration is given in Appendix~\ref{app:qualitative-example}.
 
\subsection{Dose--Response}
\label{sub:dose_response}

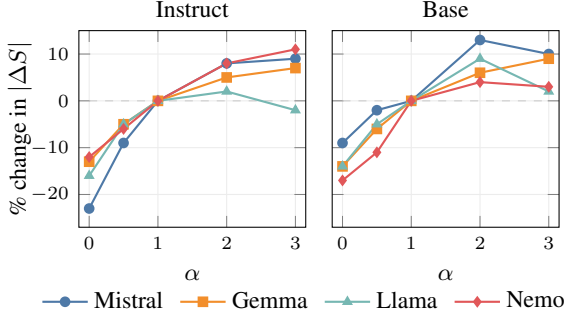
\begin{figure}[h]
\centering
\begin{tikzpicture}
\begin{axis}[
    name=plotL,
    scale only axis, width=3.0cm, height=2.6cm,
    xlabel={$\alpha$}, ylabel={\% change in $|\Delta S|$},
    title={\footnotesize Instruct}, title style={yshift=-1ex},
    xlabel style={font=\footnotesize},
    ylabel style={font=\footnotesize, yshift=-1.6ex},
    x tick label style={font=\scriptsize}, y tick label style={font=\scriptsize},
    xmin=-0.15, xmax=3.15, ymin=-27, ymax=15,
    xtick={0,1,2,3}, ytick={-20,-10,0,10},
    grid=major, grid style={gray!15},
    legend style={
        at={(1.00,-0.27)}, anchor=north,
        font=\footnotesize, draw=none, legend columns=4,
        /tikz/every even column/.append style={column sep=4pt},
    },
    every axis plot/.append style={thick, mark size=1.6pt},
]
\draw[gray!40, dashed] (axis cs:0,0) -- (axis cs:3,0);
\addplot[color=mistralC, mark=*]        coordinates {(0,-23)(0.5,-9)(1,0)(2,8)(3,9)};
\addplot[color=gemmaC,   mark=square*]  coordinates {(0,-13)(0.5,-5)(1,0)(2,5)(3,7)};
\addplot[color=llamaC,   mark=triangle*]coordinates {(0,-16)(0.5,-5)(1,0)(2,2)(3,-2)};
\addplot[color=nemoC,    mark=diamond*] coordinates {(0,-12)(0.5,-6)(1,0)(2,8)(3,11)};
\legend{Mistral, Gemma, Llama, Nemo}
\end{axis}
\begin{axis}[
    at={(plotL.east)}, anchor=west, xshift=0.35cm,
    scale only axis, width=3.0cm, height=2.6cm,
    xlabel={$\alpha$}, ylabel={},
    title={\footnotesize Base}, title style={yshift=-1ex},
    xlabel style={font=\footnotesize},
    x tick label style={font=\scriptsize}, yticklabels={},
    xmin=-0.15, xmax=3.15, ymin=-27, ymax=15,
    xtick={0,1,2,3}, ytick={-20,-10,0,10},
    grid=major, grid style={gray!15},
    every axis plot/.append style={thick, mark size=1.6pt},
]
\draw[gray!40, dashed] (axis cs:0,0) -- (axis cs:3,0);
\addplot[color=mistralC, mark=*]        coordinates {(0,-9)(0.5,-2)(1,0)(2,13)(3,10)};
\addplot[color=gemmaC,   mark=square*]  coordinates {(0,-14)(0.5,-6)(1,0)(2,6)(3,9)};
\addplot[color=llamaC,   mark=triangle*]coordinates {(0,-14)(0.5,-5)(1,0)(2,9)(3,2)};
\addplot[color=nemoC,    mark=diamond*] coordinates {(0,-17)(0.5,-11)(1,0)(2,4)(3,3)};
\end{axis}
\end{tikzpicture}
\vspace{-0.5cm}
\caption{Dose--response on instruct and base models. \% change in $|\Delta S|$ relative to baseline ($\alpha{=}1$).}
\label{fig:dose-response}
\end{figure}

Three response patterns appear across the eight models when $|\Delta S|$ is plotted as a function of $\alpha$ (details in Table \ref{tab:dose}, Appendix~\ref{app:dose-table}). (i)~Monotonic: Mistral, Gemma and Nemo Instruct, and Gemma Base. The binding strength is weakened at $\alpha = 0$ ($-12$ to $-23$\%). It strengthens above baseline at amplification ($\alpha \in [2,3]$, $+5$ to $+11$\%). (ii)~Mildly saturating at high $\alpha$: Mistral Base and Nemo Base. (iii)~Non-monotonic: Llama, where the binding effect drops substantially at $\alpha = 3$ in both versions. Llama Instruct is the only model whose binding effect crosses below baseline (Figure \ref{fig:dose-response}). 

\subsection{Generation Steering on Instruct Models} \label{gen_steering_instruct}

Table \ref{tab:steering} presents the effect of $\alpha$-scaling on the original N4 benchmark with greedy generation for the four instruct models. We add $\alpha = 5$ and drop intermediate values compared to the dose-response on logits (\S\ref{sub:dose_response}) because greedy decoding is less sensitive to small changes. This allows us to reveal the differentiation versus equal-treatment trade-off when $\alpha$ is larger. We use two accuracy metrics on~N4. $\neq_\text{acc}$ is the cultural differentiation accuracy, that is, the fraction of cultural questions where the model \rev{picks the identity that N4 labels as associated}. $=_\text{acc}$~is the equal-treatment accuracy, which represents the fraction of neutral questions answered with equal treatment. We report their changes after intervention as $\Delta\neq_\text{acc}$ and $\Delta=_\text{acc}$, in percentage points relative to the baseline $\alpha = 1$. 

At knockout ($\alpha = 0$), $\Delta\neq_\text{acc}$ is negative. Suppressing the binding heads reduces the ability of the models to differentiate 
\rev{on cultural questions}. Meanwhile, $\Delta=_\text{acc}$ stays nearly flat. When the amplification is moderate ($\alpha \in [2,3]$), the pattern reverses. $\Delta\neq_\text{acc}$ increases but $\Delta=_\text{acc}$ decreases by less than a percentage point for Mistral, Nemo and Gemma. However, the steering is not satisfactory in Llama since $\Delta=_\text{acc}$ decreases more than $\Delta\neq_\text{acc}$ increases. When the amplification is stronger ($\alpha = 5$), the dissociation starts to weaken, and while $\Delta\neq_\text{acc}$ still increases, $\Delta=_\text{acc}$ degrades strongly. Overall, when the amplification is too strong, models start differentiating 
\rev{on neutral questions as well}. Base models do not show a reliable steering trade-off at generation (Appendix~\ref{app:base_steering}), which is consistent with their smaller binding strength.

\begin{table}[h]
\centering
\small
\setlength{\tabcolsep}{3pt}
\begin{tabular}{@{}llrrrr@{}}
\toprule
$\alpha$ & & Mistral & Nemo & Llama & Gemma \\
\midrule
\multirow{2}{*}{0} & $\Delta\neq_\text{acc}$ & $-$4.5 & $-$2.2 & $-$3.5 & $-$2.3 \\
 & $\Delta=_\text{acc}$ & $+$0.1 & $+$0.1 & $+$1.2 & $+$0.3 \\
\midrule
\rowcolor[gray]{.93}
\multirow{2}{*}{1} & $\Delta\neq_\text{acc}$ & \phantom{$+$}0.0 & \phantom{$+$}0.0 & \phantom{$+$}0.0 & \phantom{$+$}0.0 \\
\rowcolor[gray]{.93}
 & $\Delta=_\text{acc}$ & \phantom{$+$}0.0 & \phantom{$+$}0.0 & \phantom{$+$}0.0 & \phantom{$+$}0.0 \\
\midrule
\multirow{2}{*}{2} & $\Delta\neq_\text{acc}$ & $+$2.6 & $+$1.6 & $+$1.3 & $+$2.1 \\
 & $\Delta=_\text{acc}$ & $-$0.4 & $-$0.6 & $-$1.0 & $-$0.3 \\
\midrule
\multirow{2}{*}{3} & $\Delta\neq_\text{acc}$ & $+$2.4 & $+$2.2 & $+$1.3 & $+$2.2 \\
 & $\Delta=_\text{acc}$ & $-$0.8 & $-$0.6 & $-$1.7 & $-$0.8 \\
\midrule
\multirow{2}{*}{5} & $\Delta\neq_\text{acc}$ & $+$3.1 & $+$4.2 & $+$2.2 & $+$3.0 \\
 & $\Delta=_\text{acc}$ & $-$2.2 & $-$2.2 & $-$5.0 & $-$1.9 \\
\bottomrule
\end{tabular}
\caption{Generation-time steering on N4 (instruct models). Changes in pp relative to $\alpha = 1$.}
\label{tab:steering}
\vspace{-0.3cm}
\end{table}

\subsection{Association--Differentiation Gap}
\label{sec:results:association-diff-gap}
The knowledge probing task (\S \ref{sec:method:knowledge_probing}) measures how much the output of a model shifts when the associated identity $R$ is present in a direct association question, outside of a decisional scenario. The same factorial pairs and formula as in the binding task are used. This allows us to compute a knowledge pairing effect $|\Delta K|$. We compare $\Delta K$ and $\Delta S$ at baseline (this section) and under knockout (\S \ref{sec:result:knockout_dissociation}). Table \ref{tab:knowledge-gap} shows the association--differentiation gap (visualized for the instruct models  in Appendix~\ref{app:knowledge-gap-fig}). Across all eight models, the knowledge pairing effect $|\Delta K|$ is much larger than the binding pairing effect $|\Delta S|$. Both pairing effects are negative and statistically significant for the eight models. Full statistics are available in Appendix \ref{app:pairing-significance}. The ratio $|\Delta K|/|\Delta S|$ is stable at 3.5$-$5$\times$ for instruct and 3.3$-$6$\times$ for base. The absolute gap is larger for instruct models than for base models, yet the ratio is comparable (Table~\ref{tab:knowledge-gap}). Models pick the related identity $R$ more often when asked directly (knowledge probe) than when the same association must drive a decision (binding). For instance, Gemma-2-9B-it chooses $R$ on $0.89$ of knowledge prompts versus $0.26$ of decision prompts. The behavioral gap between knowledge and action is large for all models (instruct: $0.22$--$0.63$, base: $0.32$--$0.56$). Details are provided in Appendix~\ref{app:behav-acc}. 

Conditioning on the cases where the model commits to an identity (argmax $\neq$ c), we further ask whether that choice is the correct $R$. On the knowledge probe, this conditional accuracy is high for all eight models ($0.90$--$0.97$). On the binding task it is also significantly above chance for every model except Gemma-2-9B-it ($0.575$, $p = 0.22$). Even when this model differentiates, it selects the contextually appropriate identity barely above chance, despite clearly knowing the association (Table~\ref{tab:behav-condacc}, Appendix~\ref{app:behav-acc}). The gap points to whether the model acts on the association, not whether it holds it.

\begin{table}[h]\centering\small
\setlength{\tabcolsep}{3pt}
\begin{tabular}{l ccc@{\hspace{10pt}}ccc}
\toprule
& \multicolumn{3}{c}{\textbf{Instruct}} & \multicolumn{3}{c}{\textbf{Base}} \\
\cmidrule(lr){2-4}\cmidrule(lr){5-7}
\textbf{Model} & $|\Delta S|$ & $|\Delta K|$ & $K/S$ & $|\Delta S|$ & $|\Delta K|$ & $K/S$ \\
\midrule
Mistral-7B & 3.45 & 12.31 & 3.57 & 0.044 & 0.262 & 5.97 \\
Gemma-2-9B & 1.95 & 8.69 & 4.47 & 0.192 & 0.955 & 4.99 \\
Llama-3.1-8B & 0.98 & 4.90 & 5.01 & 0.049 & 0.161 & 3.27 \\
Nemo-12B & 0.91 & 4.57 & 5.03 & 0.103 & 0.418 & 4.07 \\
\bottomrule
\end{tabular}
\caption{Association--differentiation gap. 
All $|\Delta S|$ and $|\Delta K|$ values are significant. Results in Appendix~\ref{app:pairing-significance}.}\label{tab:knowledge-gap}
\vspace{-0.3cm}
\end{table}

\subsection{Knockout Dissociation}
\label{sec:result:knockout_dissociation}

The knockout affects binding in two different ways, which we treat separately: its \textit{magnitude} (how strongly the model differentiates, below) and its \textit{direction} (whether it still chooses the correct identity, Appendix~\ref{app:directional-ko}). Table \ref{tab:ko-dissociation} presents the effect of the R$\rightarrow$item knockout on binding versus knowledge. We reuse the heads discovered on the instruct models (\S\ref{sub:head_discovery}, \S\ref{results:identified_heads}), since our question is whether the \textit{same} heads that mediate binding also carry knowledge, rather than where knowledge-specific heads are located. As a complementary check, we re-run the discovery pipeline independently on the knowledge prompts (using the same three attention features and $\geq 3/5$-fold stability rule). Crucially, no separate knowledge circuit emerges. In every model but Gemma-2-9B-it, the knowledge group contains the binding heads, differing only by a few knowledge-leaning heads. In Gemma-2-9B-it, one binding head (L16H15) is absent from the knowledge group. On the instruct models, knocking out these extra knowledge-leaning heads lowers $|\Delta K|$ far more than $|\Delta S|$. They contribute to retrieval rather than to the decision (Appendix~\ref{app:khead-discovery}).

\begin{table}[h!]
\centering
\footnotesize
\setlength{\tabcolsep}{2pt}
\begin{tabular}{l ccc@{\hspace{6pt}}ccc}
\toprule
& \multicolumn{3}{c}{\textbf{Instruct}} & \multicolumn{3}{c}{\textbf{Base}} \\
\cmidrule(lr){2-4}\cmidrule(lr){5-7}
& $\%\,S$ & $\%\,K$ & $K/S$ & $\%\,S$ & $\%\,K$ & $K/S$ \\
\midrule
Mistral-7B   & $-$23.5 & $-$26.6 & 1.13 & \phantom{0}$-$9.0 & $-$42.1 & 4.67 \\
Nemo-12B     & $-$12.3 & $-$12.3 & 1.00 & $-$17.4 & $-$19.0 & 1.09 \\
Llama-3.1-8B & $-$16.4 & $-$31.1 & 1.90 & $-$14.5 & $-$36.4 & 2.51 \\
Gemma-2-9B   & $-$13.2 & \phantom{0}$-$4.8 & 0.36 & $-$13.8 & $-$21.6 & 1.57 \\
\bottomrule
\end{tabular}
\caption{Change in $|\Delta|$ (\%) after R$\rightarrow$item knockout on binding ($S$) versus knowledge ($K$). $K/S$ is unitless.}
\label{tab:ko-dissociation}
\end{table}

The $K/S$ ratios vary across models. For all architectures, $K/S$ is larger for base than for instruct. Gemma-2-9B-it is the only model with $K/S < 1$, the single case where knockout affects the decision more than retrieval. $K/S$ is a ratio of relative change. Models with small $|\Delta S|$ can mechanically lead to higher ratios. We consider it indicative only.

\paragraph{U$\rightarrow$item Control} When the non-associated edges are removed, $|\Delta K|$ increases by 2.6$-$6.6\% (Appendix~\ref{app:specificity-controls}). The pattern is similar to $|\Delta S|$, whose results are reported in Table~\ref{tab:ko-summary}. The parallel effects on both tasks confirm that the knockout effect is specific to the R$\rightarrow$item edge and is not just an artifact of the attention being disrupted.

\paragraph{Directional Knockout}\label{par:directional-ko} Apart from the magnitude, knocking out R$\rightarrow$item impacts the \textit{direction} of differentiation. In the match condition, $P(R \mid \{a,b\})$ drops toward chance under R$\rightarrow$item knockout in all eight models, and the drop is statistically significant. On the other hand, the U$\rightarrow$item control does not reduce $P(R \mid \{a,b\})$ and even slightly increases it. The effect is small in base models ($|\Delta_R| \leq 0.011$, where $\Delta_R$ is the change in $P(R\mid\{a,b\})$ under R$\rightarrow$item knockout) but solid in instruct ($|\Delta_R|$ up to $0.082$ for Mistral-7B). This shows that the identified heads do not only participate in scaling the binding signal but also steer it toward the contextually appropriate identity (Appendix~\ref{app:directional-ko}). The heads are not the only carriers of this directional signal. An exploratory neuron-level directional screening procedure on the same axis identifies candidate MLP neurons aligned toward or away from $R$. Mean-ablating the toward-$R$ ones lowers $P(R\mid\{a,b\})$ on cultural prompts while leaving an independent neutral control almost flat (Appendix~\ref{app:mlp-neurons}). Combined with the binding heads, they reach $1.4$--$1.7\times$ the directional effect of the heads alone. This is sub-additive, suggesting the two may partially overlap. This holds for three of the four instruct models, not Gemma.
 
\section{Discussion}

\paragraph{What do the binding heads compute? } We observe that R$\rightarrow$item edge knockout affects binding across models, while U$\rightarrow$item knockout does not reduce it (Table~\ref{tab:ko-summary}). This clarifies what the heads we identified actually do. They likely act as a matching circuit: comparing identity and item representations, and shifting the output away from equal treatment when a match is detected. This is similar to the finding of \citet{feng2024how} who showed that LLMs use binding ID vectors to associate entities and attributes within a given context. However, in our work, the binding is not contextual but based on prior knowledge. This matching has a \textit{magnitude} as well as a \textit{direction}. Beyond modulating whether the model differentiates, the heads also steer the choice 
\rev{toward the related identity} (Appendix~\ref{app:directional-ko}). 

\paragraph{Why mid-layers?} The heads we identified cluster in mid-layers, between 20\% and 38\% of network depth (layers 7$-$16 across all four architectures, whose depth ranges from 32 layers for Mistral-7B and Llama to 42 layers for Gemma). 
Mid-layers are well-positioned to cross-reference item and identity. They are late enough that both representations are available but early enough to act before the decision is committed. This matches prior observations that factual recall in transformers tends to be mediated by mid-layers. \citet{davies2023discoveringvariablebindingcircuitry} localized their variable-binding heads within the same depth range.

\paragraph{The Llama anomaly} Llama is the outlier with a non-monotonic dose-response (Figure~\ref{fig:dose-response}), the highest $K/S$ ($\approx$ 1.9 instruct, Table~\ref{tab:ko-dissociation}), and an unfavorable steering trade-off (it loses more in $=_\text{acc}$ than it gains in $\neq_\text{acc}$ at $\alpha \in [3, 5]$, Table~\ref{tab:steering}). Its heads still carry the correct direction of differentiation (Tables~\ref{tab:behav-condacc} and~\ref{tab:directional-ko}). The anomaly is limited to amplification. A possible explanation is that they are more involved in knowledge retrieval than gating (thus the high $K/S$, consistent with its retrieval-leaning K-head, Appendix~\ref{app:khead-discovery}), so an important part of the binding signal passes through pathways our pipeline did not capture. That might explain why amplifying them injects retrieval without a matching increase in gating, leading to over-differentiation on neutral questions at high $\alpha$.

\paragraph{Instruction tuning amplifies an existing mechanism}
\label{disc:instruction_tuning_amplifies}
The instruct$\rightarrow$base transfer (\S \ref{results:identified_heads}) indicates that the cultural matching circuit is not introduced by instruction tuning but emerges during pre-training. What instruction tuning changes seems to be twofold. On one hand, it amplifies the signal. The binding strength $|\Delta S|$ values are much larger in the instruct compared to the base models (Table~\ref{tab:effect-all}), while the relative effect of the intervention is similar in both of them (Table~\ref{tab:ko-summary}). It means that the same heads participate in carrying the binding signal in the base and instruct versions, but more strongly on instruct. All four base models show a $K/S$ ratio above 1 after knockout (1.09$-$4.67, see Table~\ref{tab:ko-dissociation}). This indicates that the heads are more involved in knowledge retrieval than in gating in the base versions.
On the other hand, the $K/S$ drops in every model after instruction tuning (Table~\ref{tab:ko-dissociation}). The magnitude of the change varies (from very strong for Mistral and Gemma to modest for Llama and Nemo), but the direction is consistent. Instruction tuning seems to increase the gating role of these heads relative to knowledge retrieval. Even after this shift, the instruct balance varies across architectures ($K/S \approx 1$ for Mistral and Nemo, $\approx 1.9$ for Llama, $<1$ for Gemma, Table~\ref{tab:ko-dissociation}). Within the identified set the two roles are entangled rather than cleanly modular. We can partially separate them at the individual head level. Retrieval concentrates in a few additional heads and Gemma has one binding-specific head (Appendix~\ref{app:khead-discovery}). We treat this balance as an observation rather than a stable taxonomy.

\paragraph{Explanation of the implicit$-$explicit gap} \citet{veselovsky2025localizedculturalknowledgeconserved} qualitatively observed that models have knowledge that does not resurface without being provided explicit cultural cues. Our results provide a quantitative and mechanistic account of this finding. We find a 3.5$-$5$\times$ ratio between $|\Delta K|$ and $|\Delta S|$ in instruct models and 3.3$-$6$\times$ in base. Models know more than they differentiate. Moreover, the association--differentiation gap is present in base models too, indicating it does not arise from instruction tuning. It is not an artifact of the logit space either since it persists under behavioral argmax (Appendix~\ref{app:behav-acc}). The bottleneck lies in mobilizing the knowledge at decision time.

\section{Conclusion} \label{sec:conclusion}\label{sec:future_work} 
Our work presented a mechanistic investigation of cultural binding in LLMs
\rev{, the link between a cultural item and its related identity.} We reframed context-appropriate cultural differentiation as a binding problem rather than a bias problem, \rev{shifting the question from how to remove a behavior to how it is computed and how it can be modulated}. The 2$-$3 attention heads we identify are a small but tractable handle on that competence. The gap we measure between what models know and what they act upon suggests that difference-aware behavior is limited by gating rather than by knowledge. This is an invitation to study other normative domains, such as legal or medical reasoning, where the same distinction matters.

\section*{Limitations} \label{sec:limitations}
\paragraph{Scope of the benchmark} The N4 benchmark covers cultural appropriation with 66 items and 77 identities, which is a relatively small sample. Its identities and scenarios are mainly US-centric, so the associations we study reflect a Western perception of cultural belonging. These associations may not transfer to other cultural contexts.

\paragraph{Partial account of the mechanism} The heads are discovered based on their attention patterns (the query-key, QK side). This does not capture heads that write a strong binding signal through the output-value (OV) circuit while attending diffusely. Edge knockout is the most surgical intervention available, so the 9--23\% it removes is best read as a lower bound on the involvement of these heads. \rev{Self-repair points in the same direction. When a component is ablated, later layers can partly compensate for it \citep{mcgrath2023hydraeffectemergentselfrepair}. Our knockout measures the effect after this compensation. The direct contribution of the heads may therefore be larger.} An exploratory neuron-level analysis suggests that part of the directional signal is also carried by MLP neurons (\S\ref{sec:result:knockout_dissociation}, Appendix~\ref{app:mlp-neurons}). This contribution combines with that of the heads. The remainder (OV-circuit contributions and residual-stream directions) is left unexplained. \rev{Path patching is the natural next step: it would recover the OV-mediated part of the circuit and measure the direct contribution of the heads.} 

\paragraph{Multiple-choice setting} Both the binding pipeline (which requires locating identity and item token positions) and the knowledge probing task rely on multiple-choice questions. Such questions are also exposed to bias related to option position \citep{zheng2024large}. This is partly mitigated by randomizing $R$ across (a) and (b). Using the differences between match$-$mismatch, in which (c) is constant, also cancels position preference. However, this is exact only if that preference is additive. A residual bias could remain if the model's tendency to pick (c) depends on the content of options (a) and (b), which differs between conditions. \rev{The knowledge task uses the same option layout, so a position bias would affect $K$ and $S$ alike and cannot explain the association--differentiation gap. A full permutation control over the three positions is left to future work.}

\paragraph{Circularity between selection and testing} Our heads are selected by their ability to discriminate match from mismatch prompts. They are then causally tested on the same distinction. The knockout is not an independent confirmation. It tests whether correlationally selected heads are also causally implicated. Two observations limit the concern: random heads from the same layers do not produce a comparable effect, and the edge specificity (R$\rightarrow$item reduces binding, U$\rightarrow$item does not) was never optimized for by the L1 classifier. 


\section*{Ethics Statement}
\label{sec:ethics}
Our results have a practical consequence for how interventions on culturally aware behavior might be designed. The limiting factor is not knowledge acquisition but its mobilization at decision time. Instead of providing models with additional cultural data or retraining for cultural awareness, an alternative would be to act on the circuit mediating this mobilization. The results of our $\alpha$-scaling on three of four instruct models are a proof of concept that the identified heads participate in mediating cultural differentiation. This is not a deployable fairness method yet, in particular because the steering does not transfer cleanly to Llama and the magnitude of the gain is modest. However, this suggests that targeted and circuit-level interventions are a promising alternative to data-centric or training-centric approaches for difference-aware behavior. That would first require studying their side effects. Indeed, steering a localized circuit may shift model behavior beyond the targeted cultural differentiation.

The same mechanism has a dual use. The intervention that amplifies context-appropriate differentiation can, when applied in the opposite direction or at high $\alpha$, suppress it or push the model toward over-differentiation on neutral questions. Such a behavior would reinforce rather than mitigate cultural stereotyping. Any deployment would require calibration and auditing.

What we consider the ``correct'' identity to associate with an item is inherited from the N4 benchmark. Differentiating based on an identity and deeming it appropriate encodes a particular view of cultural belonging. The associations we steer toward cannot be taken as ground truth.

\section*{Acknowledgments}
We thank the reviewers for their constructive comments, which helped improve this paper.

\bibliography{sources}

\appendix

\section{Model Identifiers}
\label{app:model_ids}

Table~\ref{tab:model-ids} lists the Hugging Face identifiers of the eight models (instruct and base versions) used in~\S\ref{sec:experiments:setup}.

\begin{table}[h]\centering\scriptsize
\setlength{\tabcolsep}{4pt}
\begin{tabular}{@{}ll@{}}
\toprule
\textbf{Model} & \textbf{HF identifier} \\
\midrule
Mistral-7B inst. & \texttt{mistralai/Mistral-7B-Instruct-v0.3} \\
Mistral-7B base & \texttt{mistralai/Mistral-7B-v0.3} \\
Llama-3.1-8B inst. & \texttt{meta-llama/Llama-3.1-8B-Instruct} \\
Llama-3.1-8B base & \texttt{meta-llama/Llama-3.1-8B} \\
Gemma-2-9B inst. & \texttt{google/gemma-2-9b-it} \\
Gemma-2-9B base & \texttt{google/gemma-2-9b} \\
Nemo-12B inst. & \texttt{mistralai/Mistral-Nemo-Instruct-2407} \\
Nemo-12B base & \texttt{mistralai/Mistral-Nemo-Base-2407} \\
\bottomrule
\end{tabular}
\caption{Hugging Face model identifiers.}\label{tab:model-ids}
\end{table}

\section{Detailed S-score Results}
\label{app:detailed_s_score}

We report here (Table~\ref{tab:effect-all-details}) the mean $\bar{S}$ values for the match and mismatch conditions on each model, complementing the aggregated binding strength $|\Delta S|$ reported in Table~\ref{tab:effect-all} (\S\ref{sec:experiments:setup}).

\begin{table}[h]\centering\small
\setlength{\tabcolsep}{2.5pt}
\begin{tabular}{llcc|cc}
\toprule
& & \multicolumn{2}{c|}{\textbf{Instruct}} & \multicolumn{2}{c}{\textbf{Base}} \\
\textbf{Model} & \textbf{Cond.} & $\bar{S}$ & $|\Delta S|$ & $\bar{S}$ & $|\Delta S|$ \\
\midrule
\multirow{2}{*}{Mistral-7B} & Match & 2.98 & \multirow{2}{*}{\textbf{3.45}} & $-$0.568 & \multirow{2}{*}{0.044} \\
& Mismatch & 6.43 & & $-$0.524 \\
\midrule
\multirow{2}{*}{Gemma-2-9B} & Match & 0.81 & \multirow{2}{*}{\textbf{1.95}} & $-$0.851 & \multirow{2}{*}{0.192} \\
& Mismatch & 2.75 & & $-$0.660 \\
\midrule
\multirow{2}{*}{Llama-3.1-8B} & Match & $-$1.44 & \multirow{2}{*}{\textbf{0.98}} & $-$0.635 & \multirow{2}{*}{0.049} \\
& Mismatch & $-$0.46 & & $-$0.585 \\
\midrule
\multirow{2}{*}{Nemo-12B} & Match & $-$1.93 & \multirow{2}{*}{\textbf{0.91}} & $-$0.753 & \multirow{2}{*}{0.103} \\
& Mismatch & $-$1.02 & & $-$0.651 \\
\bottomrule
\end{tabular}
\caption{Detailed pairing effect on instruct and base models.}\label{tab:effect-all-details}
\end{table}

\section{Directional Preference}

\label{app:directional}

The pairing effect $\Delta S$ measures the inclination of a model to differentiate. However, it does not take into account \emph{which} of (a) or (b) is picked, as mentioned in \S\ref{sec:experiments:setup}. 

\begin{equation}
P(R \mid \{a, b\}) = \frac{P(R)}{P(a) + P(b)}
\label{eq:directional}
\end{equation}

\noindent This is the probability that the model chooses the related identity $R$, which is $P(R) = P(a)$ if $R$ is in position $(a)$, and $P(R) = P(b)$ otherwise. \newline
Table~\ref{tab:directional} reports $P(R \mid \{a,b\})$ values in the match and mismatch conditions for the eight models. Instruct models show a stronger directional preference (0.60--0.74) than base models (0.51--0.55). This suggests that instruction tuning strengthens (rather than introduces) both the inclination of the models to differentiate and the directionality of cultural differentiation.

\begin{table}[h]\centering\footnotesize
\setlength{\tabcolsep}{4pt}
\begin{tabular}{l cc c@{\hspace{6pt}} cc c}
\toprule
& \multicolumn{3}{c}{\textbf{Instruct}} & \multicolumn{3}{c}{\textbf{Base}} \\
\cmidrule(lr){2-4}\cmidrule(lr){5-7}
\textbf{Model} & Match & Mism. & $x$ & Match & Mism. & $x$ \\
\midrule
Mistral-7B    & 0.739 & 0.494 & 14 & 0.512 & 0.497 & \phantom{0}6 \\
Gemma-2-9B    & 0.599 & 0.503 & \phantom{0}2 & 0.554 & 0.503 & \phantom{0}9 \\
Llama-3.1-8B  & 0.664 & 0.495 & 11 & 0.544 & 0.499 & 13 \\
Nemo-12B      & 0.733 & 0.502 & 18 & 0.552 & 0.503 & 15 \\
\bottomrule
\end{tabular}
\caption{Directional differentiation $P(R \mid \{a,b\})$. Column $x$ gives the significance of the match value versus $0.5$ as $p<10^{-x}$ (cluster-corrected $t$-test, $n=66$).}\label{tab:directional}
\end{table}

\section{Statistical Significance of Pairing Effects}
\label{app:pairing-significance}

We provide the statistical evidence for the significance claims made in \S\ref{sec:experiments:setup} and \S\ref{sec:results:association-diff-gap}, where we state that the pairing effects $\Delta S$ and $\Delta K$ differ from zero across all eight models.

To do so, we use a cluster-corrected t-test against zero. Among the 847 factorial pairs we use, many share an item with other pairs, violating the independence assumption. We aggregate the per-pair $\Delta$ values into item-level means. We have $n=66$ independent observations, that is, one per cultural item. 
Table~\ref{tab:pairing-significance} shows the $t$-statistics for all models. 

\begin{table}[h]\centering\small
\setlength{\tabcolsep}{3pt}
\begin{tabular}{l cc cc}
\toprule
& \multicolumn{2}{c}{\textbf{Instruct}} & \multicolumn{2}{c}{\textbf{Base}} \\
\cmidrule(lr){2-3}\cmidrule(lr){4-5}
\textbf{Model} & $t_{\Delta S}$ & $t_{\Delta K}$ & $t_{\Delta S}$ & $t_{\Delta K}$ \\
\midrule
Mistral-7B   & $-6.97$ & $-14.35$ & $-5.32$ & $-11.76$ \\
Gemma-2-9B   & $-5.31$ & $-16.70$ & $-8.11$ & $-13.60$ \\
Llama-3.1-8B & $-6.29$ & $-13.84$ & $-5.03$ & $-9.87$  \\
Nemo-12B     & $-5.43$ & $-15.31$ & $-6.04$ & $-12.36$ \\
\bottomrule
\end{tabular}
\caption{Cluster-corrected $t$-test against zero on item-level means ($n=66$). All $p < 0.001$.}\label{tab:pairing-significance}
\end{table}

\section{Detailed Edge-Knockout}
\label{app:detailed_edge_knockout}

We extend Table~\ref{tab:ko-summary} here (Table~\ref{tab:ko-all-detail}) from \S\ref{results:identified_heads} by reporting the $|\Delta S|$ values before and after knockout, in addition to the percentage changes.

\begin{table}[H]\centering\small
\setlength{\tabcolsep}{3pt}
\begin{tabular}{lcc|cc}
\toprule
& \multicolumn{2}{c|}{\textbf{Instruct}} & \multicolumn{2}{c}{\textbf{Base}} \\
\textbf{Model} & $|\Delta S|$ & \textbf{Change} & $|\Delta S|$ & \textbf{Change} \\
\midrule
\multicolumn{5}{l}{\emph{Baseline (no knockout)}} \\
Mistral-7B & 3.45 & --- & 0.0438 & --- \\
Gemma-2-9B & 1.95 & --- & 0.1915 & --- \\
Llama-3.1-8B & 0.98 & --- & 0.0491 & --- \\
Nemo-12B & 0.91 & --- & 0.1026 & --- \\
\midrule
\multicolumn{5}{l}{\emph{R$\to$item knockout}} \\
Mistral-7B & 2.64 & $-$23.5\% & 0.0399 & \phantom{0}$-$9.0\% \\
Gemma-2-9B & 1.69 & $-$13.2\% & 0.1651 & $-$13.8\% \\
Llama-3.1-8B & 0.82 & $-$16.4\% & 0.0420 & $-$14.5\% \\
Nemo-12B & 0.80 & $-$12.3\% & 0.0848 & $-$17.4\% \\
\midrule
\multicolumn{5}{l}{\emph{U$\to$item knockout (control)}} \\
Mistral-7B & 3.65 & +5.8\% & 0.0468 & +6.7\% \\
Gemma-2-9B & 2.05 & +5.6\% & 0.2026 & +5.8\% \\
Llama-3.1-8B & 1.04 & +5.9\% & 0.0549 & +11.8\% \\
Nemo-12B & 0.96 & +5.3\% & 0.1021 & $-$0.5\% \\
\bottomrule
\end{tabular}
\caption{Full edge-knockout results across eight models.}\label{tab:ko-all-detail}
\end{table}

\section{Specificity Controls}
\label{app:specificity-controls}

The two specificity controls discussed in \S\ref{sub:random_baseline} (binding) and \S\ref{sec:result:knockout_dissociation} (knowledge) are reported here. The random-head baseline (Table~\ref{tab:random-baseline}) is run on the instruct models, where the binding signal is largest. None of the 400 random samples reaches the identified-head effect. We did not run the random-head control on the base models. Their $|\Delta S|$ (Table~\ref{tab:effect-all}) is much smaller and would make the comparison less sensitive. The second control, the U$\rightarrow$item knockout (Table~\ref{tab:a-control-knowledge}), shows only a small increase in $|\Delta K|$. This confirms edge specificity.

\begin{table}[H]\centering\footnotesize
\setlength{\tabcolsep}{1pt}
\begin{tabular}{@{}c@{\hspace{3pt}}l c c c c@{}}
\toprule
& & \textbf{Ident.} & \multicolumn{2}{c}{\textbf{Random} (n=100)} & \textbf{Emp.} \\
\cmidrule(lr){4-5}
& \textbf{Model} & change & mean$\pm$std & 5th & $p$ \\
\midrule
\multirow{4}{*}{\rotatebox[origin=c]{90}{\textit{Instruct}}}
& Mistral-7B   & $-$23.5\% & $-0.2{\pm}1.0$\% & $-1.8$\% & $<$0.01 \\
& Gemma-2-9B   & $-$13.2\% & $+0.2{\pm}1.0$\% & $-0.9$\% & $<$0.01 \\
& Llama-3.1-8B & $-$16.4\% & $\phantom{+}0.0{\pm}0.7$\% & $-0.9$\% & $<$0.01 \\
& Nemo-12B     & $-$12.3\% & $-0.2{\pm}0.8$\% & $-1.2$\% & $<$0.01 \\
\bottomrule
\end{tabular}
\caption{Random head knockout baseline on instruct models. Empirical $p$ is one-sided.}
\label{tab:random-baseline}
\end{table}

\begin{table}[H]\centering\small
\setlength{\tabcolsep}{5pt}
\begin{tabular}{l cc}
\toprule
& \multicolumn{2}{c}{\textbf{$|\Delta K|$}} \\
\cmidrule(lr){2-3}
\textbf{Model} & Instruct & Base \\
\midrule
Mistral-7B & $+$6.6\% & $+$4.4\% \\
Nemo-12B & $+$3.1\% & $+$3.8\% \\
Llama-3.1-8B & $+$6.2\% & $+$4.5\% \\
Gemma-2-9B & $+$2.6\% & $+$3.8\% \\
\bottomrule
\end{tabular}
\caption{U$\rightarrow$item control knockout: \% change in $|\Delta K|$.}\label{tab:a-control-knowledge}
\end{table}

\section{Item-Level Knockout Example}
\label{app:qualitative-example}

We illustrate the R$\rightarrow$item knockout mentioned in~\S\ref{sub:random_baseline} at the item level, using the five factorial pairs that cover the Holi festival in Mistral-7B-Instruct. At baseline, $|\Delta S| = 6.00$, indicating strong differentiation when the associated identity is present. Knocking out the R$\rightarrow$item edges of the three identified heads (L8H16, L9H23, L12H9) lowers $|\Delta S|$ to 3.95, a 34.2\% reduction at the item level (larger than the 23.5\% average across all 66 items in Table~\ref{tab:ko-summary}).

\section{Dose--Response Values}
\label{app:dose-table}

Figure~\ref{fig:dose-response} (in the main body of text) shows the curves for both instruct and base models, and Table~\ref{tab:dose} gives the dose--response values for the eight models (\S\ref{sub:dose_response}). Cells are shaded by the behavior between $\alpha = 2$ and $\alpha = 3$. \colorbox{cr}{Red} shows a non-monotonic reversal (the effect at $\alpha = 3$ falls below its $\alpha = 2$ value by more than 5~pp, or crosses below baseline). \colorbox{cy}{Yellow} illustrates mild saturation.

\begin{table}[H]\centering\small
\setlength{\tabcolsep}{2pt}
\definecolor{cg}{HTML}{C8E6C9}
\definecolor{cy}{HTML}{FFF3C4}
\definecolor{cr}{HTML}{FFCDD2}
\begin{tabular}{l cc@{\hspace{10pt}}cc@{\hspace{10pt}}cc@{\hspace{10pt}}cc}
\toprule
& \multicolumn{2}{c}{\textbf{Mistral}} & \multicolumn{2}{c}{\textbf{Gemma}} & \multicolumn{2}{c}{\textbf{Llama}} & \multicolumn{2}{c}{\textbf{Nemo}} \\
\cmidrule(lr){2-3}\cmidrule(lr){4-5}\cmidrule(lr){6-7}\cmidrule(lr){8-9}
$\boldsymbol{\alpha}$ & Inst. & Base & Inst. & Base & Inst. & Base & Inst. & Base \\
\midrule
\phantom{0.}0 & $-$23 & \phantom{0}$-$9 & $-$13 & $-$14 & $-$16 & $-$14 & $-$12 & $-$17 \\
0.5 & \phantom{0}$-$9 & \phantom{0}$-$2 & \phantom{0}$-$5 & \phantom{0}$-$6 & \phantom{0}$-$5 & \phantom{0}$-$5 & \phantom{0}$-$6 & $-$11 \\
\rowcolor[gray]{.93}
\phantom{0.}1 & \phantom{+0}0 & \phantom{+0}0 & \phantom{+0}0 & \phantom{+0}0 & \phantom{+0}0 & \phantom{+0}0 & \phantom{+0}0 & \phantom{+0}0 \\
\phantom{0.}2 & \phantom{0}+8 & +13 & \phantom{0}+5 & \phantom{0}+6 & \phantom{0}+2 & \phantom{0}+9 & \phantom{0}+8 & \phantom{0}+4 \\
\phantom{0.}3 & \phantom{0}+9 & \cellcolor{cy}+10 & \phantom{0}+7 & \phantom{0}+9 & \cellcolor{cr}\phantom{0}$-$2 & \cellcolor{cr}\phantom{0}+2 & +11 & \cellcolor{cy}\phantom{0}+3 \\
\bottomrule
\end{tabular}
\caption{Dose--response: \% change in binding strength $|\Delta S|$ relative to baseline ($\alpha{=}1$).}\label{tab:dose}
\end{table}

\section{Base Model Steering Results}
\label{app:base_steering}

Table~\ref{tab:steering-base} reports the steering results on base models, which are mentioned in \S\ref{gen_steering_instruct}. 
\begin{table}[h!]
\centering
\small
\setlength{\tabcolsep}{3pt}
\begin{tabular}{@{}llrrrr@{}}
\toprule
$\alpha$ & & Mistral & Nemo & Llama & Gemma \\
\midrule
\multirow{2}{*}{0} & $\Delta\neq_\text{acc}$ & $+$2.2 & $-$0.9 & $-$1.6 & $-$5.0 \\
 & $\Delta=_\text{acc}$ & $+$0.9 & $+$0.4 & $+$1.2 & $+$1.1 \\
\midrule
\rowcolor[gray]{.93}
\multirow{2}{*}{1} & $\Delta\neq_\text{acc}$ & \phantom{$+$}0.0 & \phantom{$+$}0.0 & \phantom{$+$}0.0 & \phantom{$+$}0.0 \\
\rowcolor[gray]{.93}
 & $\Delta=_\text{acc}$ & \phantom{$+$}0.0 & \phantom{$+$}0.0 & \phantom{$+$}0.0 & \phantom{$+$}0.0 \\
\midrule
\multirow{2}{*}{2} & $\Delta\neq_\text{acc}$ & $+$0.0 & $+$0.2 & $-$0.1 & $+$0.0 \\
 & $\Delta=_\text{acc}$ & $+$0.5 & $+$1.5 & $+$1.3 & $-$0.4 \\
\midrule
\multirow{2}{*}{3} & $\Delta\neq_\text{acc}$ & $-$0.6 & $-$0.1 & $+$0.1 & $+$0.6 \\
 & $\Delta=_\text{acc}$ & $+$0.6 & $+$0.1 & $+$1.9 & $-$1.6 \\
\midrule
\multirow{2}{*}{5} & $\Delta\neq_\text{acc}$ & $-$0.3 & $+$1.0 & $+$0.6 & $+$1.8 \\
 & $\Delta=_\text{acc}$ & $-$1.4 & $-$1.8 & $+$1.8 & $-$2.8 \\
\bottomrule
\end{tabular}
\caption{Generation-time steering on N4 (base models). Changes in pp relative to $\alpha = 1$.}
\label{tab:steering-base}
\end{table}

\section{Visualization of the Association--Differentiation Gap}
\label{app:knowledge-gap-fig}

\begin{figure}[h]
\centering
\begin{tikzpicture}
\begin{axis}[
    ybar,
    area legend,
    width=0.92\columnwidth,
    height=5cm,
    bar width=7pt,
    ylabel={Pairing effect $|\Delta|$},
    ylabel style={font=\small},
    symbolic x coords={Mistral, Gemma, Llama, Nemo},
    xtick=data,
    x tick label style={font=\small},
    y tick label style={font=\small},
    ymin=0, ymax=15,
    legend style={
        at={(0.5,-0.15)}, anchor=north,
        font=\footnotesize, draw=none,
        row sep=1pt, column sep=5pt,
    },
    enlarge x limits=0.2,
    grid=major,
    grid style={gray!15},
]
\addplot[fill=Scolor!80, draw=Scolor!80] coordinates {
    (Mistral, 3.45) (Gemma, 1.95) (Llama, 0.98) (Nemo, 0.91)
};
\addplot[fill=Kcolor!80, draw=Kcolor!80] coordinates {
    (Mistral, 12.31) (Gemma, 8.69) (Llama, 4.90) (Nemo, 4.57)
};
\legend{$|\Delta S|$ (binding), $|\Delta K|$ (knowledge)}
\node[font=\tiny, text=black!60] at (axis cs:Mistral, 13.2) {$3.6\times$};
\node[font=\tiny, text=black!60] at (axis cs:Gemma, 9.6) {$4.5\times$};
\node[font=\tiny, text=black!60] at (axis cs:Llama, 5.8) {$5.0\times$};
\node[font=\tiny, text=black!60] at (axis cs:Nemo, 5.5) {$5.0\times$};
\end{axis}
\end{tikzpicture}
\caption{Association--differentiation gap on instruct models.}
\label{fig:knowledge-gap}
\end{figure}
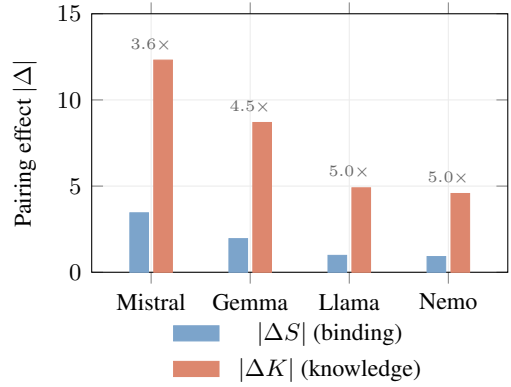

Figure~\ref{fig:knowledge-gap} plots the association--differentiation gap, which was mentioned in Table~\ref{tab:knowledge-gap} (\S\ref{sec:results:association-diff-gap}). The knowledge pairing effect $|\Delta K|$ is about 3.5--5$\times$ larger than the binding pairing effect $|\Delta S|$ across all instruct models.

\section{Behavioral Differentiation Accuracy}
\label{app:behav-acc}

The directional preference of Appendix~\ref{app:directional} operates at the logits level while this appendix details the behavioral knowledge-to-action gap reported in \S\ref{sec:results:association-diff-gap}. We introduce a behavioral measure of differentiation by using the argmax over the three option tokens $(\ell_a, \ell_b, \ell_c)$. For each match prompt, we observe whether the model picks $R$, $U$ or the non-differentiation option~$(c)$. For mismatch prompts, we record whether it correctly picks~$(c)$. All values are item-level means~($n=66$). 

Table~\ref{tab:behav-acc} shows that the association is expressed more strongly in the knowledge probe than in the decision one, leaving a behavioral knowledge-to-action gap of $0.22$--$0.63$. The gap is also present in base models. It is consistent with the logit results that the association--differentiation gap exists before instruction tuning. The mismatch column illustrates how often the model correctly picks ``neither'' when no identity is associated. This is lower and more variable across models than the match-side preference for $R$. The knowledge-to-action gap is therefore driven by the match condition (the model recognizes the related identity but does not act on it) rather than by the abstention behavior.

\begin{table}[h]\centering\footnotesize
\setlength{\tabcolsep}{2pt}
\begin{tabular}{c l cccc}
\toprule
& \textbf{Model} & Bind.\ $R$ & Know.\ $R$ & Gap & Know.\ $(c)_\text{mism}$ \\
\midrule
\multirow{4}{*}{\rotatebox[origin=c]{90}{\textit{Instruct}}}
& Mistral-7B   & 0.267 & 0.853 & 0.585 & 0.542 \\
& Gemma-2-9B   & 0.260 & 0.888 & 0.628 & 0.682 \\
& Llama-3.1-8B & 0.606 & 0.838 & 0.232 & 0.729 \\
& Nemo-12B     & 0.710 & 0.932 & 0.223 & 0.380 \\
\midrule
\multirow{4}{*}{\rotatebox[origin=c]{90}{\textit{Base}}}
& Mistral-7B   & 0.250 & 0.809 & 0.560 & 0.644 \\
& Gemma-2-9B   & 0.593 & 0.910 & 0.318 & 0.611 \\
& Llama-3.1-8B & 0.439 & 0.801 & 0.363 & 0.523 \\
& Nemo-12B     & 0.482 & 0.944 & 0.462 & 0.259 \\
\bottomrule
\end{tabular}
\caption{Behavioral differentiation accuracy (argmax, item-level means). Bind.\ $R$ and Know.\ $R$ are the fraction of match prompts where the model picks the related identity in the binding versus the knowledge task. Gap $=$ Know.\ $R$ $-$ Bind.\ $R$. Know.\ $(c)_\text{mism}$ is the fraction of knowledge mismatch prompts where the model correctly picks ``neither''.}
\label{tab:behav-acc}
\end{table}

In a second step, we condition on the cases where the model commits to an identity (argmax $\neq$ c). Table~\ref{tab:behav-condacc} reports the accuracy of that choice. That is, how often the committed identity is $R$. On the knowledge probe, this conditional accuracy is high and significantly above chance for all eight models ($0.90$--$0.97$, all $p < 0.001$). On the binding task, it is also high and significant for all models except Gemma-2-9B-it, whose value does not differ from chance ($0.575$, $p = 0.22$). This is consistent with its low $K/S$ ratio and large knowledge-to-action gap: even when this model commits to an identity, its choice is close to random in the decision context, although it knows the association.

\begin{table}[h]\centering\footnotesize
\setlength{\tabcolsep}{5pt}
\begin{tabular}{@{}c@{\hspace{5pt}}lcc@{}}
\toprule
& \textbf{Model} & Binding & Knowledge \\
\midrule
\multirow{4}{*}{\rotatebox[origin=c]{90}{\textit{Instruct}}}
& Mistral-7B   & $0.859^{***}$ & $0.974^{***}$ \\
& Gemma-2-9B   & $0.575^{\,\text{n.s.}}$ & $0.954^{***}$ \\
& Llama-3.1-8B & $0.743^{***}$ & $0.908^{***}$ \\
& Nemo-12B     & $0.827^{***}$ & $0.934^{***}$ \\
\midrule
\multirow{4}{*}{\rotatebox[origin=c]{90}{\textit{Base}}}
& Mistral-7B   & $0.720^{***}$ & $0.926^{***}$ \\
& Gemma-2-9B   & $0.757^{***}$ & $0.941^{***}$ \\
& Llama-3.1-8B & $0.830^{***}$ & $0.899^{***}$ \\
& Nemo-12B     & $0.863^{***}$ & $0.946^{***}$ \\
\bottomrule
\end{tabular}
\caption{Conditional differentiation accuracy: $P(\text{argmax}=R \mid \text{argmax} \neq c)$ in the match condition (item-level means, $n=66$). Significance from $t$-test against $0.5$ ($^{***}p<0.001$, n.s.\ $p>0.05$).}
\label{tab:behav-condacc}
\end{table}

\section{Directional Preference under Knockout} 
\label{app:directional-ko}

Appendix~\ref{app:directional} shows that, at baseline, models prefer the related identity $R$ once they differentiate. This appendix supports the directional knockout result discussed in \S\ref{sec:result:knockout_dissociation} by extending the directional analysis to the knockout conditions. We recompute the same directional preference $P(R \mid \{a,b\})$ under knockout, to test whether the identified heads carry the \emph{direction} of that preference (toward $R$) and not only its magnitude (Table~\ref{tab:ko-dissociation}). Table~\ref{tab:directional-ko} presents the results. Ablating the R$\rightarrow$item edge reduces $P(R \mid \{a,b\})$ toward chance (0.5) in every model ($\Delta_R < 0$, all $p < 0.001$). It lowers the preference for the related identity but does not eliminate it. This is consistent with the identified heads carrying only part of the binding mechanism. As a control, ablating the U$\rightarrow$item edge (the non-associated identity) never reduces $P(R)$, and even slightly increases it ($\Delta_U \geq 0$). Only the edge toward the associated identity carries the directional preference. The effect is small in base models (whose baseline preference is close to chance), but still statistically significant and it is much larger in instruct models.

\begin{table}[h]\centering\footnotesize
\setlength{\tabcolsep}{5pt}
\begin{tabular}{@{}c@{\hspace{5pt}}lccc@{}}
\toprule
& \textbf{Model} & $P_0$ & $\Delta_R$ & $\Delta_U$ \\
\midrule
\multirow{4}{*}{\rotatebox[origin=c]{90}{\textit{Instruct}}}
& Mistral-7B   & 0.739 & $-0.082^{***}$ & $+0.010^{***}$ \\
& Gemma-2-9B   & 0.599 & $-0.016^{***}$ & $+0.002^{\,\text{n.s.}}$ \\
& Llama-3.1-8B & 0.664 & $-0.041^{***}$ & $+0.008^{***}$ \\
& Nemo-12B     & 0.733 & $-0.032^{***}$ & $+0.008^{***}$ \\
\midrule
\multirow{4}{*}{\rotatebox[origin=c]{90}{\textit{Base}}}
& Mistral-7B   & 0.512 & $-0.004^{***}$ & $+0.000^{\,\text{n.s.}}$ \\
& Gemma-2-9B   & 0.554 & $-0.010^{***}$ & $+0.003^{***}$ \\
& Llama-3.1-8B & 0.544 & $-0.011^{***}$ & $+0.003^{***}$ \\
& Nemo-12B     & 0.552 & $-0.008^{***}$ & $+0.001^{\,\text{n.s.}}$ \\
\bottomrule
\end{tabular}
\caption{Directional effect of the knockout on $P(R\mid\{a,b\})$ (match condition). $\Delta_R = P(R)_{\text{R-KO}} - P_0$, $\Delta_U = P(R)_{\text{U-KO}} - P_0$. $P_0$ is the pre-KO baseline. Paired $t$-test on item-level means, $n=66$, $^{***}$ is $p<10^{-3}$.}
\label{tab:directional-ko}
\end{table}

\section{Independent Head Discovery on the Knowledge Task}
\label{app:khead-discovery}

The dissociation of \S\ref{sec:result:knockout_dissociation} reuses the binding heads. As a complementary check, we re-run the discovery pipeline from \S\ref{sub:head_discovery}
on the knowledge prompts, then knock out the resulting heads on both the binding and knowledge tasks for instruct models. Table~\ref{tab:khead-groups} shows the retained group. In every model but Gemma-2-9B-it, the knowledge group contains the binding heads: no separate set of knowledge heads emerges. Where it differs, it adds knowledge-leaning heads that lower $|\Delta K|$ far more than $|\Delta S|$. Conversely, in Gemma-2-9B-it, one of the binding heads (L16H15) lowers $|\Delta S|$ but not $|\Delta K|$.

\begin{table}[h]\centering\scriptsize
\setlength{\tabcolsep}{3pt}
\setlength{\fboxsep}{1.5pt}
\begin{tabular}{@{}c@{\hspace{3pt}}l@{\hspace{4pt}}l@{}}
\toprule
& \textbf{Model} & \textbf{Heads} \\
\midrule
\multirow{11}{*}{\rotatebox[origin=c]{90}{\textit{Instruct}}}
& \multirow{2}{*}{Mistral-7B}
  & B: L8H16, L9H23, L12H9 \\
& & K: L8H16, L9H23, L12H9 \\
\addlinespace[2pt]
& \multirow{2}{*}{Gemma-2-9B}
  & B: L11H14, L13H13, \colorbox{cr}{L16H15} \\
& & K: L11H14, L13H13 \\
\addlinespace[2pt]
& \multirow{2}{*}{Llama-3.1-8B}
  & B: L7H7, L8H17 \\
& & K: L7H7, L8H17, \colorbox{cy}{L11H14} \\
\addlinespace[2pt]
& \multirow{2}{*}{Nemo-12B}
  & B: L8H9, L10H24 \\
& & K: L8H9, L10H24, \colorbox{cy}{L12H15} \\
\cmidrule(l){2-3}
\multirow{11}{*}{\rotatebox[origin=c]{90}{\textit{Base}}}
& \multirow{2}{*}{Mistral-7B}
  & B: L8H16, L9H23, L12H9 \\
& & K: L8H16, L9H23, L12H9 \\
\addlinespace[2pt]
& \multirow{2}{*}{Gemma-2-9B}
  & B: L11H14, L13H13, L16H15 \\
& & K: L11H14, L13H13, L16H15, \colorbox{cy}{L16H14} \\
\addlinespace[2pt]
& \multirow{2}{*}{Llama-3.1-8B}
  & B: L7H7, L8H17 \\
& & K: L7H7, L8H17, \colorbox{cy}{L11H14} \\
\addlinespace[2pt]
& \multirow{2}{*}{Nemo-12B}
  & B: L8H9, L10H24 \\
& & K: L8H9, L10H24, \colorbox{cy}{L15H28} \\
\bottomrule
\end{tabular}
\setlength{\fboxsep}{3pt}
\caption{Binding ($B$) versus Knowledge ($K$) group per model. Heads in normal text are shared between the two groups. \colorbox{cy}{Knowledge-leaning} heads appear only in $K$. The \colorbox{cr}{binding-only} head (Gemma-2-9B-it, L16H15) appears only in $B$.}
\label{tab:khead-groups}
\end{table}

\noindent The group knockout confirms this (Tables~\ref{tab:khead-group-ko}, \ref{tab:khead-base-K}). 
Adding a knowledge-leaning head raises the knowledge reduction (Llama $-31.1\%\to-43.5\%$) far more than the binding reduction ($-16.4\%\to-19.8\%$). The added head is more knowledge-skewed (incremental ratio $\approx 3.6$) than the binding group it is added to ($\approx 1.9$). This is why we read it as retrieval-leaning. Conversely, for the binding-only head, on Gemma-2-9B-it, adding L16H15 to \{L11H14, L13H13\} reduces the binding from $-10.1\%$ to $-13.2\%$ while the knowledge reduction barely moves (from $-4.5\%$ to $-4.8\%$). 

On the base models, we only report $\Delta K$, because the base binding signal is small (0.04--0.19, Table~\ref{tab:effect-all}), so the incremental effect of one added head on $|\Delta S|$ is too small to be statistically reliable.

\begin{table}[h]\centering\footnotesize
\setlength{\tabcolsep}{3pt}
\begin{tabular}{@{}c@{\hspace{5pt}}l cc@{\hspace{8pt}} l cc@{}}
\toprule
& & \multicolumn{2}{c}{\textbf{Binding group}} & \multicolumn{3}{c}{\textbf{+\,K-leaning head}} \\
\cmidrule(lr){3-4}\cmidrule(lr){5-7}
& \textbf{Model} & $\Delta S$ & $\Delta K$ & head & $\Delta S$ & $\Delta K$ \\
\midrule
\multirow{4}{*}{\rotatebox[origin=c]{90}{\textit{Instruct}}}
& Mistral-7B   & $-$23.5 & $-$26.6 & ---    & ---     & ---     \\
& Gemma-2-9B   & $-$13.2 & \phantom{0}$-$4.8 & ---    & ---     & ---     \\
& Llama-3.1-8B & $-$16.4 & $-$31.1 & L11H14 & $-$19.8 & $-$43.5 \\
& Nemo-12B     & $-$12.3 & $-$12.3 & L12H15 & $-$12.8 & $-$14.9 \\
\bottomrule
\end{tabular}
\caption{Group R$\rightarrow$item knockout on the instruct models, \% change in $|\Delta|$. All reported values are significant ($p<0.001$, item-level, $n=66$).}
\label{tab:khead-group-ko}
\end{table}

\begin{table}[h]\centering\footnotesize
\setlength{\tabcolsep}{5pt}
\begin{tabular}{@{}c@{\hspace{5pt}}l c l c@{}}
\toprule
& & \textbf{Binding group} & \multicolumn{2}{c}{\textbf{+\,K-leaning head}} \\
\cmidrule(lr){3-3}\cmidrule(lr){4-5}
& \textbf{Model} & $\Delta K$ & head & $\Delta K$ \\
\midrule
\multirow{4}{*}{\rotatebox[origin=c]{90}{\textit{Base}}}
& Mistral-7B   & $-$42.1 & ---    & ---     \\
& Gemma-2-9B   & $-$21.6 & L16H14 & $-$28.0 \\
& Llama-3.1-8B & $-$36.4 & L11H14 & $-$48.8 \\
& Nemo-12B     & $-$19.0 & L15H28 & $-$21.4 \\
\bottomrule
\end{tabular}
\caption{Group R$\rightarrow$item knockout on the base models, \% change in $|\Delta K|$ ($\Delta S$ omitted). All values are significant ($p<0.001$, item-level, $n=66$).}
\label{tab:khead-base-K}
\end{table}

\section{MLP Directional Screening and Head--MLP Composition}
\label{app:mlp-neurons}

In \S\ref{sec:result:knockout_dissociation} and Appendix~\ref{app:directional-ko}, the preference toward $R$ is primarily associated with the binding heads. Here, we perform an exploratory analysis to test whether MLP neurons selected using a directional screening procedure also affect the same preference and how the two interact. Both experiments run on the instruct models. We measure the directional preference $P(R\mid\{a,b\})$ in the match condition (item-level means, $n=66$).

\paragraph{Neuron-level directional screening} MLP neurons write vectors into the residual stream through their down-projection column. We project this column onto $u_{\text{dir}}$. This is the unembedding direction separating the logits of the two identity options and oriented toward $R$. For each neuron, we compute a directional proxy $\Delta_{\text{dir}}$, a neuron-level direct logit attribution (DLA). It is then scaled by the neuron's activation and normalized by the residual RMS. We seek a linear proxy to estimate how strongly the neuron's output aligns with the $R$-versus-$U$ decision direction. A neuron may also align with the same direction on \textit{neutral} prompts, that is, when there is no identity that should be favored. This is what we call a leak. We rank neurons by a specificity ratio $|\Delta_{\text{dir}}|/(\text{leak}+\epsilon)$ to keep the cleanest $200$. This score is only used for candidate selection and should not be interpreted as a causal attribution. We assess causal relevance separately, through mean-ablation. We mean-ablate the top-$K$ toward-$R$ (\textsc{pos}, $\Delta_{\text{dir}}>0$) or anti-$R$ (\textsc{neg}, $\Delta_{\text{dir}}<0$) neurons. The leak is measured on half the neutral prompts (split A) and the post-ablation control is run on the other half (split B). The control is never evaluated on the data used for the selection. As Table~\ref{tab:mlp-ksweep} shows, suppressing neurons selected by the directional screening procedure lowers $P(R)$ on cultural prompts. The neutral control only moves a few points at most. The effect is two-sided for Llama, one-sided (\textsc{pos}) for Mistral and Nemo, and absent for Gemma.

\begin{table}[h]\centering\footnotesize
\setlength{\tabcolsep}{3pt}
\begin{tabular}{@{}c@{\hspace{5pt}}l l ccc@{}}
\toprule
& \textbf{Model} & & $K{=}10$ & $K{=}20$ & $K{=}40$ \\
\midrule
\multirow{8}{*}{\rotatebox[origin=c]{90}{\textit{Instruct}}}
& \multirow{2}{*}{Mistral-7B}
& \textsc{pos} & $+0.002$    & $-0.030^{*}$ & $-0.078^{*}$ \\
& & \textsc{neg} & $-0.001$    & $-0.000$     & $+0.003^{*}$ \\
\addlinespace[2pt]
& \multirow{2}{*}{Llama-3.1-8B}
& \textsc{pos} & $-0.015^{*}$ & $-0.023^{*}$ & $-0.032^{*}$ \\
& & \textsc{neg} & $+0.010^{*}$ & $+0.012^{*}$ & $+0.029^{*}$ \\
\addlinespace[2pt]
& \multirow{2}{*}{Nemo-12B}
& \textsc{pos} & $-0.010^{*}$ & $-0.025^{*}$ & $-0.033^{*}$ \\
& & \textsc{neg} & $+0.004$     & $+0.007^{*}$ & $-0.010$     \\
\addlinespace[2pt]
& \multirow{2}{*}{Gemma-2-9B}
& \textsc{pos} & $+0.001$     & $+0.003$     & $-0.006$     \\
& & \textsc{neg} & $-0.001$     & $+0.008^{*}$ & $+0.009$     \\
\bottomrule
\end{tabular}\caption{Directional screening and MLP-neuron ablation (instruct). $\Delta P(R\mid\{a,b\})$ in the match condition after mean-ablating the top-$K$ toward-$R$ (\textsc{pos}) or anti-$R$ (\textsc{neg}) neurons. $^{*}$: $p<0.05$ (item-level $t$-test, $n=66$). }
\label{tab:mlp-ksweep} 
\end{table}

\paragraph{Head--MLP additivity} We run three ablations on $P(R)$: heads only (\textbf{A}, R$\rightarrow$item edges, reproducing $\Delta_R$ of Table~\ref{tab:directional-ko}), top-$K$ toward-$R$ neurons only (\textbf{B}, $K=40$ for Mistral, $20$ for Llama and Nemo), and both together (\textbf{C}). Table~\ref{tab:mlp-additivity} shows that the combined effect $C$ is smaller in magnitude than the sum of the individual effects, $|C| < |A+B|$ ($C/(A{+}B)=0.87$--$0.92$). The result is significant for Llama and Nemo and marginal for Mistral. This suggests that the heads and the selected neurons are not completely independent and may partially rely on overlapping pathways. Together, they produce a larger directional effect than the head-only intervention, reaching $1.4$--$1.7\times$ its magnitude. We did not include Gemma since it has no significant directional MLP component.

\begin{table}[h]\centering\footnotesize
\setlength{\tabcolsep}{4pt}
\begin{tabular}{@{}c@{\hspace{5pt}}l cccc@{}}
\toprule
& \textbf{Model} & \textbf{A} & \textbf{B} & \textbf{C} & $C/(A{+}B)$ \\
\midrule
\multirow{3}{*}{\rotatebox[origin=c]{90}{\textit{Instruct}}}
& Mistral-7B   & $-0.082$ & $-0.079$ & $-0.141$ & $0.87^{\dagger}$ \\
& Llama-3.1-8B & $-0.041$ & $-0.023$ & $-0.057$ & $0.89^{***}$ \\
& Nemo-12B     & $-0.032$ & $-0.026$ & $-0.054$ & $0.92^{***}$ \\
\bottomrule
\end{tabular}
\caption{Head--MLP additivity on $P(R\mid\{a,b\})$ in the match condition (instruct, $n=66$). $^{\dagger}p=0.05$, $^{***}p<0.001$, paired $t$-test.}
\label{tab:mlp-additivity}
\end{table}


\end{document}